\documentclass[conference]{IEEEtran}
\IEEEoverridecommandlockouts
\usepackage{cite}
\usepackage{amsmath,amssymb,amsfonts}
\usepackage{algorithmic}
\usepackage{graphicx}
\usepackage{textcomp}
\usepackage{xcolor}
\usepackage{multirow}
\usepackage{tikz}
\usepackage{booktabs}
\usepackage{array}
\usepackage{tabularx}
\usepackage{makecell}
\usepackage{enumitem}
\usepackage{siunitx}
\usepackage{url}
\usepackage[hidelinks]{hyperref}
\usetikzlibrary{arrows.meta,positioning,fit,calc}
\def\BibTeX{{\rm B\kern-.05em{\sc i\kern-.025em b}\kern-.08em
    T\kern-.1667em\lower.7ex\hbox{E}\kern-.125emX}}
\begin{document}

\title{REPORT-OPS: Hybrid Evidence-Guided Reporting with Human-Supervised Refinement for HPC
\thanks{\IEEEauthorrefmark{1} Konate and Akewar contributed equally to this work.}

}

\author{
\IEEEauthorblockN{
Mah Kadidia Konate\IEEEauthorrefmark{2}\IEEEauthorrefmark{1},
Mayur Akewar\IEEEauthorrefmark{3}\IEEEauthorrefmark{1},
Janki Bhimani\IEEEauthorrefmark{3},
Steven Farrell\IEEEauthorrefmark{2}}
\IEEEauthorblockA{\IEEEauthorrefmark{2}\textit{NERSC, Lawrence Berkeley National Laboratory, Berkeley, California, USA}}
\IEEEauthorblockA{\IEEEauthorrefmark{3}\textit{Florida International University, Miami, Florida, USA}}
\IEEEauthorblockA{\texttt{kadidiakonate@lbl.gov, makew001@fiu.edu, jbhimani@fiu.edu, sfarrell@lbl.gov}}
}

\maketitle
\thispagestyle{plain}
\pagestyle{plain}
\begin{abstract}
Operational reporting for high-performance computing (HPC) facilities requires combining information from monitoring services, schedulers, incident and maintenance records, storage measurements, periodic reports, and operator investigations. Producing a reliable report requires more than retrieving relevant records: the evidence must refer to the correct system, metric definition, time period, and source of authority. We present \textsc{REPORT-OPS}, a hybrid reporting agent in which an LLM interprets requests, plans evidence collection, evaluates possible explanations, and drafts reports, while author-seeded operational policies and automated checks determine whether the available evidence supports the requested conclusion. Depending on the evidence state, the system answers the request, reports a qualified conclusion, or withholds an unsupported claim. During offline refinement, an LLM critic proposes bounded policy changes that a human approves or rejects without editing; accepted candidates are promoted only after automated target, held-out, and regression testing.

We evaluate \textsc{REPORT-OPS} on 68 sanitized reporting queries constructed from 50 base scenarios in a frozen HPC operational corpus, using three agent models and two LLM judges. \textsc{REPORT-OPS} achieves 68.63\% Pass@80, compared with 7.02\% for the strongest retrieval baseline. Evidence gating increases claim-support precision from 40.58\% to 90.06\%. On the 13-query explanatory subset, explanation accuracy reaches 55.54\%. Across five rounds of human-supervised refinement, localized policy updates correct 91.17\% of targeted failures and improve performance on related held-out cases by 9.33 percentage points, with no observed regression breaks under release gating. These results demonstrate improved evidence-grounded reporting and within-environment policy reuse; they do not establish unrestricted causal discovery or cross-facility generalization.
\end{abstract}


\section{Introduction}
\label{sec:introduction}

High-performance computing (HPC) facilities collect operational evidence from monitoring systems, schedulers, storage services, incident records, support systems, operator discussions, and periodic reports. These sources support reporting on utilization, availability, queue behavior, workload activity, storage performance, reliability, and user support. Producing a trustworthy report, however, requires more than retrieving relevant records or summarizing measurements. Each claim must be interpreted using the correct metric definition, system entity, reporting period, source authority, and relationship to other operational conditions. Production studies demonstrate substantial variation in workloads, resource usage, scheduling behavior, failures, and I/O activity across systems and time periods \cite{yang2024lifecycle,amvrosiadis2018diversity}. Scientific user facilities have therefore identified provenance, transparency, human oversight, and safe integration as important requirements for operational AI systems \cite{hellert2026agentic, 318638, 11575426, 10.1145/3736548.3737835}.

Consider a request asking why utilization on Perlmutter, NERSC’s flagship supercomputer, decreased during a particular month. Utilization alone cannot distinguish among several possible causes: reduced demand, scheduler draining, large-job placement constraints, reservations, unavailable nodes, maintenance, outages, or incomplete accounting data. Resolving this requires evidence, but not all evidence is equally usable. A deep eligible-job queue, for instance, contradicts any claim of reduced demand, since sustained queuing indicates continued load. An outage record is relevant only if it overlaps both the affected system and the reporting interval; an outage on a different system, or outside the window, is not evidence at all. Operator discussion can offer useful diagnostic context, but it is not authoritative for official metrics. The agent must therefore judge not only which evidence is relevant, but whether it is sufficient, compatible, timely, and authoritative for the specific conclusion at hand. This is a \emph{reporting-fitness} problem: evidence useful for one claim may be unfit for another, because sources differ in authority, semantics, and temporal scope.

Large language models (LLMs) provide useful foundations for accessing complex HPC information. \textsc{ChatHPC}, \textsc{AskHPC}, and \textsc{ATHOS} demonstrate question answering, retrieval, tool integration, and configurable evaluation for HPC environments \cite{yin2025chathpc,bondapalli2025askhpc,farrell2026athos}. Retrieval-augmented generation and tool-using reasoning enable models to access external records and operational systems \cite{lewis2020rag,yao2023react,chen2024rcacopilot}, but relevance alone does not establish reporting validity: a retrieved record may concern the wrong system or period, use an incompatible definition, or lack authority for the requested conclusion. Process-oriented evaluation and provenance methods improve traceability \cite{he2025trajectbench,kim2025trace,wang2026provenance,alvarez2026provenanceguard}, but do not generally define the claim-specific evidence that must be present before an HPC report should answer, qualify, or block.

A second challenge is that operational explanations depend on relationships across evidence types, not just the validity of each item in isolation. For example, a spike in queue wait time might plausibly be attributed to reduced capacity, yet the same signature can arise from a scheduling policy that limits how many jobs per user can accrue priority at a time, especially when many similar jobs are submitted at once; in that case, the backlog reflects a structural eligibility effect rather than any loss of availability or demand. Treating wait time as evidence in isolation, without checking it against job submission patterns and scheduling policy, can therefore produce explanations that are individually plausible but inconsistent with the broader operational picture. A third challenge concerns system improvement. Feedback-based methods can improve agent behavior \cite{shinn2023reflexion,madaan2023selfrefine,yuksekgonul2025textgrad}, but an unrestricted prompt or policy update may correct one failure while damaging previously correct behavior. Safe refinement requires localizing the responsible policy element and testing a bounded change against related and previously successful cases before release \cite{zhang2026agentdevel}.

We present \textsc{REPORT-OPS}, a hybrid agent for evidence-grounded operational reporting with offline, regression-gated policy refinement. The LLM interprets each request, plans evidence collection, proposes and tests possible explanations, links claims to retrieved records, and drafts the report. Author-seeded, versioned policies define acceptable source roles, evidence requirements, operational relations, and completion conditions. Automated checks enforce entity, metric, unit, temporal, contradiction, and evidence-validity constraints. Based on these checks, the system answers the request, qualifies the conclusion, or withholds an unsupported claim.

This separation also defines the role of human oversight. Humans do not review or edit routine online reports. During offline refinement, an LLM critic proposes a bounded change to the operational policy. A human may approve or reject the proposed patch without rewriting it, after which automated target, held-out, and regression tests determine whether it is promoted. Thus, the LLM provides flexible interpretation and generation, while explicit policy and automated verification control reporting validity and policy release.

We evaluate \textsc{REPORT-OPS} on 68 sanitized HPC reporting queries expanded from 50 base scenarios in a frozen operational corpus. The queries represent complete, missing, temporally invalid, source-inappropriate, contradictory, and unavailable evidence conditions. The evaluation uses three agent models, two independent judge models, deterministic claim- and evidence-level measures, controlled changes to operational evidence, and five rounds of human-supervised policy refinement. We investigate three research questions:

\begin{enumerate}[label=\textbf{RQ\arabic*:},leftmargin=*,itemsep=0.35em]
    \item \textbf{Evidence-guided reporting.} Can claim-specific evidence controls improve reporting reliability when evidence is complete, missing, invalid, contradictory, or unavailable?

    \item \textbf{Operational explanation.} Do structured evidence and explicit relations across workload and system conditions improve evidence-supported explanations?

    \item \textbf{Policy refinement.} Can bounded, agent-generated policy updates correct observed failures and improve related held-out cases without breaking previously correct behavior?
\end{enumerate}

This paper makes three contributions:

\begin{enumerate}[label=(\roman*),itemsep=0.25em]
    \item \textbf{A hybrid evidence-guided reporting architecture} that separates LLM-based interpretation, planning, explanation, and composition from author-seeded operational policy and automated evidence validation. This architecture makes unsupported, incomplete, and contradictory evidence visible before the system decides to answer, qualify, or withhold a conclusion.

    \item \textbf{An evidence-grounded approach to operational explanation} that connects typed workload, queue, scheduler, incident, maintenance, storage, and data-quality records. Controlled evidence changes test whether an explanation responds appropriately to relevant operational conditions; the method evaluates represented explanation families rather than claiming unrestricted causal discovery.

    \item \textbf{A human-supervised policy-refinement process} in which an LLM critic proposes bounded, trace-localized policy patches, humans approve or reject the unchanged proposals, and automated target, held-out, and regression tests control promotion.
\end{enumerate}

\section{Background and Motivation}
\label{sec:background}

\subsection{Operational Evidence and Source Roles}
\label{sec:operational_evidence}

HPC facilities observe their systems through sources with different purposes,
authority, temporal granularity, and review levels. Monitoring and scheduler
services expose current utilization, queues, node state, reservations, and
workloads; incident and maintenance records document disruptions and impact
intervals; specialized measurements characterize storage and data quality;
operator discussions preserve intermediate hypotheses and corrections; and
periodic reports provide reviewed values and staff-authored explanations.

The corpus used here reflects these roles. DOE metrics reports provide official
monthly values and commentary, live queue summaries provide short-term demand
and workload signals, utilization discussions preserve metric definitions and
data corrections, storage streams provide Community File System (CFS) baselines and tail outliers, and
incident records provide event timelines. These sources are not a single
relevance-ranked corpus. A monthly report may be authoritative for an official
utilization value but too coarse for a short anomaly; a scheduler snapshot may
establish current queue pressure but not a historical average; and unresolved
operator commentary cannot replace a verified metric or incident record. A source call returning no matching data establishes only that no fitting record was found, not that the event did not occur.

\subsection{Reporting Fitness and Operational Context}
\label{sec:reporting_fitness}

We define \emph{reporting fitness} as the suitability of an artifact for
supporting a particular report claim. Fitness requires agreement between the
claim and the artifact's source role, system entity, metric definition, unit,
observation interval, and data-health state. An artifact can therefore be
topically relevant but unfit for the requested conclusion. For example, an
otherwise correct utilization value may be invalid if it refers to a
different partition, uses a maintenance-adjusted definition when the report
requires the official definition, or covers a different reporting period.

Reporting fitness is claim-specific rather than an absolute property of a source. A scheduler
snapshot may be mandatory for a claim about current queue pressure, contextual
for a monthly utilization explanation, and irrelevant to an approved
historical value. Conversely, reviewed commentary may support a monthly
explanation while remaining insufficient for a claim about the present
system state. REPORT-OPS represents these distinctions through claim-specific
evidence requirements that identify mandatory evidence, acceptable source
roles, contextual signals, temporal conditions, and exclusions.

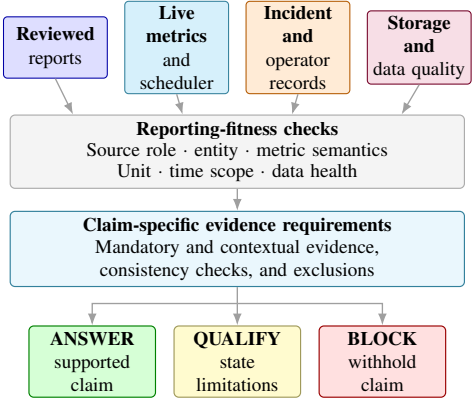
\begin{figure}[t]
\centering
\resizebox{0.70\columnwidth}{!}{%
\begin{tikzpicture}[
  >=Latex,
  source/.style={
    draw,
    rounded corners=2pt,
    align=center,
    text width=1.72cm,
    minimum height=1.12cm,
    inner sep=2.5pt,
    line width=0.55pt
  },
  stage/.style={
    draw,
    rounded corners=2.5pt,
    align=center,
    text width=8.05cm,
    minimum height=0.72cm,
    inner sep=3.5pt,
    line width=0.6pt
  },
  state/.style={
    draw,
    rounded corners=2.5pt,
    align=center,
    text width=2.10cm,
    minimum height=0.65cm,
    inner sep=3pt,
    line width=0.6pt
  },
  flow/.style={
    ->,
    line width=0.7pt,
    draw=gray!75
  }
]

\node[
  source,
  fill=blue!13,
  draw=blue!65!black
] (reports)
  {\textbf{Reviewed}\\reports};

\node[
  source,
  fill=cyan!14,
  draw=cyan!55!black,
  right=3mm of reports
] (live)
  {\textbf{Live metrics}\\and scheduler};

\node[
  source,
  fill=orange!16,
  draw=orange!70!black,
  right=3mm of live
] (records)
  {\textbf{Incident and}\\operator records};

\node[
  source,
  fill=purple!13,
  draw=purple!60!black,
  right=3mm of records
] (special)
  {\textbf{Storage and}\\data quality};

\coordinate (sourcecenter) at
  ($(reports.south)!0.5!(special.south)$);

\node[
  stage,
  fill=gray!9,
  draw=gray!65,
  below=6mm of sourcecenter
] (fitness)
  {\textbf{Reporting-fitness checks}\\[-1pt]
   Source role \(\cdot\) entity \(\cdot\) metric semantics\\[-1pt]
   Unit \(\cdot\) time scope \(\cdot\) data health};

\node[
  stage,
  fill=cyan!7,
  draw=cyan!55!black,
  below=4mm of fitness
] (requirements)
  {\textbf{Claim-specific evidence requirements}\\[-1pt]
   Mandatory and contextual evidence, consistency checks, and exclusions};

\node[
  state,
  fill=green!16,
  draw=green!50!black,
  below=7mm of requirements,
  xshift=-2.65cm
] (answer)
  {\textbf{ANSWER}\\supported claim};

\node[
  state,
  fill=yellow!23,
  draw=yellow!55!black,
  below=7mm of requirements
] (qualify)
  {\textbf{QUALIFY}\\state limitations};

\node[
  state,
  fill=red!12,
  draw=red!60!black,
  below=7mm of requirements,
  xshift=2.65cm
] (block)
  {\textbf{BLOCK}\\withhold claim};

\draw[flow]
  (reports.south) --
  ([xshift=-3.0cm]fitness.north);

\draw[flow]
  (live.south) --
  ([xshift=-1.0cm]fitness.north);

\draw[flow]
  (records.south) --
  ([xshift=1.0cm]fitness.north);

\draw[flow]
  (special.south) --
  ([xshift=3.0cm]fitness.north);

\draw[flow]
  (fitness.south) --
  (requirements.north);

\coordinate (split) at
  ($(requirements.south)+(0,-3mm)$);

\draw[
  line width=0.7pt,
  draw=gray!75
] (requirements.south) -- (split);

\draw[flow]
  (split) -| (answer.north);

\draw[flow]
  (split) -- (qualify.north);

\draw[flow]
  (split) -| (block.north);

\end{tikzpicture}%
}

\caption{Reporting fitness converts heterogeneous operational artifacts into
claim-specific completion decisions. Topical relevance alone is insufficient:
the source role, semantic scope, time scope, and data health must match the
requested claim.}
\label{fig:reporting-fitness}
\vspace{-0.3cm}
\end{figure}

Operational explanations additionally require relationships among multiple signals, rather than any single artifact in isolation. As illustrated in Section~\ref{sec:introduction}, utilization alone does not identify demand: a shallow eligible queue is consistent with reduced demand, while a deep queue points toward scheduler draining, placement constraints, reservations, resource fragmentation, unavailable nodes, or maintenance instead. Queue depth itself must be interpreted using workload size, quality-of-service constraints, account concentration, and job eligibility, since a large reported backlog does not imply that every queued job can immediately use the available nodes.

Storage measurements have similar ambiguity. A normal median read or write
rate can coexist with severe high-percentile latency affecting a subset of
I/O-intensive workloads. A short outlier may indicate transient congestion,
whereas repeated deviations from a historical baseline may indicate sustained
degradation. Storage evidence becomes relevant to an application slowdown
only when the affected filesystem, workload characteristics, and observation
interval align. Metric anomalies can also originate from the reporting
pipeline itself, including missing records, inconsistent definitions,
incorrect denominators, or delayed data imports.

Consequently, workload, queue, scheduler, incident, maintenance, storage, and
data-quality observations should be treated as evidence requirements for
testing candidate explanations, not as unconditional causal rules. Their
purpose is to distinguish supported explanations from plausible alternatives
and to expose evidence that remains missing. The resulting evidence state
determines whether REPORT-OPS should answer normally, provide a qualified
conclusion that states the limitation, or block the unsupported claim, as
shown in Figure~\ref{fig:reporting-fitness}. These decisions remain dependent
on the completeness and correctness of the instantiated requirements; a
deterministic check cannot compensate for an omitted evidence family or an
incorrectly interpreted reporting scope.

REPORT-OPS therefore performs evidence-grounded hypothesis testing rather than
unrestricted causal discovery. The LLM may propose alternatives, but the
verifier can confirm an explanation only when its evidence conditions and
operational relations are represented. An unmodeled cause should yield a
qualified or blocked conclusion, not a confident substitute from a known
explanation family.

\subsection{Related Work and Design Gap}
\label{sec:related_work}

Retrieval-augmented generation and tool-using agents ground responses in
external records and actions \cite{lewis2020rag,yao2023react}. In HPC,
\textsc{ChatHPC}, \textsc{AskHPC}, and \textsc{ATHOS} demonstrate domain
assistance, retrieval, tool integration, and configurable evaluation
\cite{yin2025chathpc,bondapalli2025askhpc,farrell2026athos}, while
\textsc{RCACopilot} demonstrates evidence collection for incident diagnosis
\cite{chen2024rcacopilot}. These systems establish the value of retrieval and
tool use, but do not generally define the authority, semantic, and temporal
conditions under which an operational reporting claim is fit for use. REPORT-OPS
therefore makes evidence requirements part of planning, retrieval, and
completion rather than relying only on post-generation checking.

Process-oriented benchmarks evaluate tool selection and reasoning trajectories
\cite{he2025trajectbench,kim2025trace}, while provenance methods connect claims
to execution records \cite{wang2026provenance}. Source-aware verification can
detect incorrect attribution after generation \cite{alvarez2026provenanceguard};
REPORT-OPS instead uses source-role requirements during planning and retrieval,
then exposes missing, invalid, contradictory, or unavailable evidence before
composition.

Feedback methods improve answers, prompts, and agent components through textual
critique~\cite{shinn2023reflexion,madaan2023selfrefine,yuksekgonul2025textgrad}.
Broader self-evolving-agent research considers changes to models, memory, tools,
prompts, and architectures~\cite{gao2026selfevolving}, while
\textsc{AgentDevel} frames agent improvement as a regression-aware release
process~\cite{zhang2026agentdevel}. \textsc{REPORT-OPS} specializes this
perspective to typed HPC reporting semantics. Candidate patches are restricted
to evidence requirements, source policies, ontology and temporal rules,
workload relations, normalization, and claim verification. A human approves or
rejects each unchanged proposal before automated target, held-out, regression,
and safety evaluation.

Regression-aware refinement is one part of the broader problem of trustworthy
foundation-model deployment. At the model level, selective-unlearning methods
study how information can be removed or modified in language models, vision
models, and embodied vision--language--action agents while limiting unintended
behavioral changes~\cite{ranjan2026forgetting,ranjan2026razor,ranjan2026vla}.
Complementary work examines privacy leakage through membership-inference
auditing~\cite{ranjan2026g}, fairness through retrieval-assisted structural
debiasing~\cite{ranjan2026catrag}. These approaches govern aspects of what
a model retains, exposes, or learns.

At the agent-system level, \textsc{SafeCommit} studies whether a
memory-grounded agent has sufficient support to perform an
action~\cite{akewar2026safecommit}. \textsc{REPORT-OPS} addresses a
complementary reporting-policy problem. It does not modify the underlying model
weights; instead, it controls which operational evidence may support a claim,
when a conclusion must be qualified or withheld, and which policy changes may
be released. Source-role validation, claim-specific evidence requirements,
human acceptance, and regression-gated promotion therefore extend
trustworthiness from model behavior to auditable operational reporting.

These gaps motivate three requirements addressed in Section~\ref{sec:design}:
claim-specific evidence must guide retrieval before generation; explanations
must connect workloads with queue, scheduler, incident, maintenance, storage,
and data-quality evidence under compatible scopes; and policy updates must be
bounded, versioned, human-supervised, and regression-gated.

\section{REPORT-OPS Design}
\label{sec:design}

\subsection{Architecture and Control Boundary}
\label{sec:overview}

\textsc{REPORT-OPS} separates \emph{online reporting} from \emph{offline policy refinement}. Each online request executes against a frozen policy version \(P_k\). This policy contains the operational ontology, evidence-requirement templates, source-role and fallback rules, workload relations, and claim-verification rules. The LLM and online tools may retrieve evidence and produce an execution trace, but they cannot modify \(P_k\). Policy changes occur only through the offline refinement process, where candidate updates are versioned, evaluated, and either promoted or rejected.

The online workflow performs eight traceable operations organized into four phases: (1) interpret the request and construct its evidence requirements; (2) plan source use, select tools, and retrieve records; (3) normalize evidence and test candidate explanations; and (4) verify claims and compose the report. This architecture is intentionally hybrid. The LLM supplies flexible interpretation, planning, hypothesis generation, and language production, while author-seeded policy defines the admissible operational semantics and automated components enforce them. Table~\ref{tab:hybrid-footprint} summarizes this division of responsibility.

\begin{table}[!hbt]
\centering
\caption{Division of responsibility in \textsc{REPORT-OPS}.}
\label{tab:hybrid-footprint}
\scriptsize
\setlength{\tabcolsep}{2.5pt}
\renewcommand{\arraystretch}{0.96}
\begin{tabularx}{\columnwidth}{@{}p{0.24\columnwidth}X@{}}
\toprule
\textbf{Element} & \textbf{Role and ownership} \\
\midrule
LLM agent &
Interprets requests, plans retrieval, proposes explanations, links claims to records, and drafts reports. \\

Author-seeded policy &
Defines ontology terms, evidence requirements, acceptable source roles, fallbacks, workload relations, and completion conditions; it is versioned and inspectable. \\

Automated verifier &
Checks schema, entity, metric, unit, temporal scope, source authority, contradictions, and evidence state before permitting a claim. \\

Offline refinement &
An LLM proposes a bounded policy patch; a human approves or rejects the unchanged proposal, and automated target, held-out, and regression tests control promotion. \\
\bottomrule
\end{tabularx}
\end{table}

Humans do not inspect or edit routine online reports. Their role is limited to the offline update boundary, where they review the safety and integrity of an agent-generated patch. They do not rewrite an accepted patch or override the automated release tests. Likewise, the deterministic verifier does not generate explanations or supply missing evidence; it determines whether proposed claims satisfy the active policy.

The policy is represented as parameterized artifact families rather than as a flat collection of query-specific rules. For example, one evidence template can instantiate different entity, source, temporal, and completion constraints for multiple requests. Workload relations define testable evidence conditions for represented explanation families rather than storing complete answers. Each policy version records its parent, modified artifacts, applicability conditions, evaluation results, and rollback target.

The control architecture is reusable beyond a particular HPC facility, but its policy content is not assumed to transfer unchanged. A deployment for another HPC site or a distributed system would retain the same interpretation, retrieval, verification, and release boundaries while replacing or extending its source adapters, entity and metric definitions, authority mappings, evidence templates, and operational relations. The evaluation in this paper tests one HPC configuration and does not claim cross-facility or non-HPC generalization.

The ablation ladder in Section~\ref{sec:configurations} separately introduces retrieval, typed representation, workload relations, evidence gating, and claim verification. These comparisons expose where explicit domain structure affects performance; they do not imply that the LLM independently learned the author-seeded reporting policy or that the evaluation fully separates every manually authored rule from the automated enforcement mechanism.

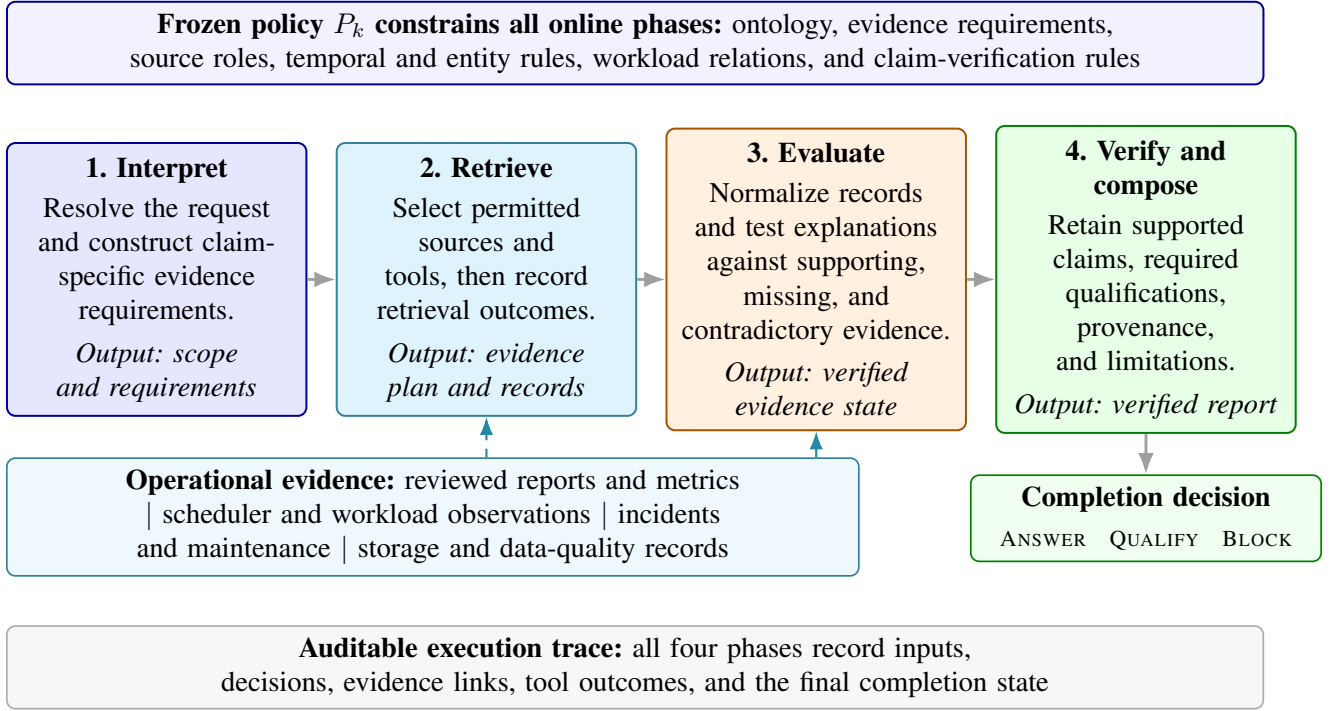
\begin{figure*}[!t]
\centering
\resizebox{0.96\textwidth}{!}{%
\begin{tikzpicture}[
>=Latex,
stage/.style={
    draw,
    rounded corners=3pt,
    align=center,
    text width=3.35cm,
    minimum height=1.42cm,
    inner sep=5pt,
    line width=0.7pt
},
policybar/.style={
    draw=blue!55!black,
    fill=blue!5,
    rounded corners=3pt,
    align=center,
    text width=15.25cm,
    minimum height=0.70cm,
    inner sep=4pt,
    line width=0.7pt
},
evidencebar/.style={
    draw=cyan!55!black,
    fill=cyan!6,
    rounded corners=3pt,
    align=center,
    text width=10.25cm,
    minimum height=0.95cm,
    inner sep=4pt,
    line width=0.6pt
},
outcomebar/.style={
    draw=green!50!black,
    fill=green!6,
    rounded corners=3pt,
    align=center,
    text width=4.05cm,
    minimum height=0.95cm,
    inner sep=4pt,
    line width=0.6pt
},
tracebar/.style={
    draw=gray!65,
    fill=gray!6,
    rounded corners=3pt,
    align=center,
    text width=15.25cm,
    minimum height=0.72cm,
    inner sep=4pt,
    line width=0.6pt
},
mainflow/.style={
    ->,
    line width=1.0pt,
    draw=gray!75
},
evidenceflow/.style={
    ->,
    dashed,
    line width=0.75pt,
    draw=cyan!60!black
}
]

\node[policybar] (policy) {
    \textbf{Frozen policy \(P_k\) constrains all online phases:}
    ontology, evidence requirements, source roles, temporal and entity rules,
    workload relations, and claim-verification rules
};

\node[
    stage,
    fill=blue!10,
    draw=blue!60!black,
    below=7mm of policy.south west,
    anchor=north west
] (interpret) {
    \textbf{1. Interpret}\\[2pt]
    Resolve the request and construct claim-specific evidence requirements.\\[3pt]
    \emph{Output: scope and requirements}
};

\node[
    stage,
    fill=cyan!11,
    draw=cyan!55!black,
    right=3.5mm of interpret
] (retrieve) {
    \textbf{2. Retrieve}\\[2pt]
    Select permitted sources and tools, then record retrieval outcomes.\\[3pt]
    \emph{Output: evidence plan and records}
};

\node[
    stage,
    fill=orange!12,
    draw=orange!65!black,
    right=3.5mm of retrieve
] (evaluate) {
    \textbf{3. Evaluate}\\[2pt]
    Normalize records and test explanations against supporting, missing, and contradictory evidence.\\[3pt]
    \emph{Output: verified evidence state}
};

\node[
    stage,
    fill=green!10,
    draw=green!50!black,
    right=3.5mm of evaluate
] (compose) {
    \textbf{4. Verify and compose}\\[2pt]
    Retain supported claims, required qualifications, provenance, and limitations.\\[3pt]
    \emph{Output: verified report}
};

\draw[mainflow] (interpret.east) -- (retrieve.west);
\draw[mainflow] (retrieve.east) -- (evaluate.west);
\draw[mainflow] (evaluate.east) -- (compose.west);

\node[
    evidencebar,
    below=5mm of interpret.south west,
    anchor=north west
] (evidence) {
    \textbf{Operational evidence:}
    reviewed reports and metrics
    \(\mid\) scheduler and workload observations
    \(\mid\) incidents and maintenance
    \(\mid\) storage and data-quality records
};

\draw[evidenceflow]
    (evidence.north -| retrieve.south) -- (retrieve.south);

\draw[evidenceflow]
    (evidence.north -| evaluate.south) -- (evaluate.south);

\node[
    outcomebar,
    below=5mm of compose
] (outcome) {
    \textbf{Completion decision}\\[2pt]
{\footnotesize
\textsc{Answer}\quad
\textsc{Qualify}\quad
\textsc{Block}}
};

\draw[mainflow] (compose.south) -- (outcome.north);

\node[
    tracebar,
    below=6mm of evidence.south west,
    anchor=north west
] (trace) {
    \textbf{Auditable execution trace:}
all four phases record inputs,\\ decisions, evidence links,
tool outcomes, and the final completion state
};

\end{tikzpicture}%
}
\caption{\textsc{REPORT-OPS} online workflow under frozen policy
\(P_k\). The agent interprets the request, retrieves operational evidence,
evaluates candidate explanations, and verifies the report before selecting
\textsc{Answer}, \textsc{Qualify}, or \textsc{Block}. Every phase contributes
to an auditable trace. Human review is outside the online path and occurs only
when an offline policy patch is proposed.}
\label{fig:online-REPORT-OPS}
\end{figure*}

\subsection{Scope and Claim-Specific Evidence Requirements}
\label{sec:evidence_requirements}

REPORT-OPS represents a request as scope
\(s(q)=\langle i,e,m,t,t_c,a\rangle\), containing its intent, entity, metric or
operational concept, observation interval, optional comparison interval, and
answer type. Temporal expressions are normalized to explicit intervals with
their timezone and intended reporting granularity. A field is inherited from
conversation history only for a valid follow-up, and the originating turn
remains in the trace. Otherwise, REPORT-OPS requests clarification.

From \(s(q)\) and \(P_k\), REPORT-OPS instantiates
\(\mathcal{E}(q)=\langle E_{\rm req},E_{\rm ctx},A,F,\Theta,X,\Pi\rangle\).
Here, \(E_{\rm req}\) and \(E_{\rm ctx}\) identify mandatory and contextual
evidence; \(A\) permits source roles; \(F\) sets freshness and coverage;
\(\Theta\) specifies entity, definition, unit, temporal, and data-health
constraints; \(X\) defines contradictions and exclusions; and \(\Pi\) defines
completion behavior. This specification states what must be established, not
what the answer should say. A metric lookup may require one authoritative value
with its definition, unit, entity, and period, whereas an explanatory request about
the same metric also requires workload, queue, scheduler, incident, maintenance,
storage, or data-quality evidence.

Each evidence slot is \textsc{Present}, \textsc{Missing}, \textsc{Invalid},
\textsc{Conflicting}, or \textsc{Unavailable}. \textsc{Present} requires at
least one record satisfying every applicable constraint; \textsc{Missing}
means that an accessible source yielded no fitting record; \textsc{Invalid}
denotes topically relevant but unfit evidence; \textsc{Conflicting} denotes
fit records that support incompatible values or conclusions; and
\textsc{Unavailable} records a permission error, timeout, source outage, or
tool failure and is never treated as negative evidence. Slot states yield three claim-level decisions:
\textsc{Answer} when the requested claim is supported, \textsc{Qualify} when a
supported subset can be reported with limitations, and \textsc{Block} when the
central value or explanation would require unsupported inference. Consequently,
REPORT-OPS may report an authoritative utilization value while blocking its cause
when workload evidence is unavailable.

\subsection{Evidence-Guided Planning, Retrieval, and Normalization}
\label{sec:retrieval_normalization}

A source capability graph maps each evidence slot to tools and sources using
evidence type, authority, entity and temporal coverage, access feasibility,
latency, and permitted fallback relations. Historical official values prefer
reviewed reports or validated metric services; current-state requests prefer
live monitoring or scheduler services. Staff discussions may contextualize an
explanation but cannot replace an authoritative historical value unless the
active requirement explicitly permits that role. Each plan entry associates an
evidence slot \(e_i\) with tool \(\tau_j\), source \(\sigma_j\), and normalized
arguments \(\phi_j\), recording both the call and why it was selected.
Independent calls execute concurrently when they have no data dependency.

Tool outcomes distinguish successful, empty, permission-denied, unavailable,
timed-out, malformed, and failed responses. A fallback satisfies a mandatory
slot only when its source role is allowed; otherwise, it remains contextual
evidence. Retrieved artifacts become typed \emph{Metric}, \emph{Queue},
\emph{Workload}, \emph{Incident}, \emph{Maintenance}, \emph{Storage},
\emph{Commentary}, or \emph{DataQuality} records. Each record retains its
identifier, entity, metric or event, value and unit, observation or impact
interval, source role, provenance, data-health state, and extraction confidence.

A record \(r\) satisfies slot \(e_i\) only when its source role, entity, metric
definition, unit, time scope, and data health are all compatible. These
deterministic checks perform timezone and unit normalization, interval
containment or overlap, metric-definition and system alignment, duplicate
detection, and conflict detection. Canonical concepts improve matching without
erasing important differences: two utilization records may share an ontology
concept while retaining different denominators, aggregation windows, or
maintenance adjustments. An invalid record remains auditable but cannot support
a claim.

\subsection{Operational Ontology and Evidence Graphs}
\label{sec:ontology_graphs}

The ontology connects user language, evidence requirements, source
capabilities, typed records, operational relations, and report claims. It
defines entities such as systems, partitions, filesystems, jobs, accounts, and
services; observations such as utilization, wait time, throughput, and node
state; events such as maintenance, outages, draining, workload campaigns, and
data corrections; and source roles such as authoritative, live, diagnostic,
incident, workload, storage, and data-quality evidence. Grounding resolves
aliases before source selection while preserving source-specific definitions
and provenance.

REPORT-OPS maintains three graph views.
The \emph{source capability graph} determines where required evidence can be
obtained. The \emph{operational evidence graph} links records by entity
identity, temporal overlap, metric compatibility, workload association, event
impact, and data quality. The \emph{claim--evidence graph} links candidate
statements to supporting, qualifying, contradicting, invalid, unavailable, or
missing evidence and therefore controls report inclusion. The LLM may propose
aliases, hypotheses, and edges, but deterministic checks require every
referenced record to exist and pass the applicable source, semantic, temporal,
and data-health checks; the LLM cannot create evidence that was not retrieved.

\subsection{Workload-Aware Explanation and Claim Verification}
\label{sec:workload_reasoning}

For explanatory requests, a versioned relation library constrains rather than
replaces LLM reasoning. Each candidate hypothesis specifies mandatory and
contextual evidence, entity and temporal predicates, contradictions, and
ranking features. For example, \emph{reduced demand} requires low eligible
demand and exclusion of stronger outage, maintenance, scheduling, and
data-quality explanations. A \emph{scheduling or placement constraint} requires
eligible queued work plus draining, reservations, job-shape incompatibility,
fragmentation, quality-of-service restrictions, or unavailable nodes.
\emph{Outage-related backlog loss} requires a disruption followed by reduced
runnable work or changed queue composition. \emph{Storage impact} requires an
abnormal storage signal, compatible filesystem and workload scope, and temporal
overlap. \emph{Measurement artifact} requires missing records, delayed imports,
definition changes, denominator errors, or documented corrections.

A hypothesis is valid only when its mandatory evidence is fit, entity and time
predicates hold, its source roles are permitted, and no blocking contradiction
is established. Only valid hypotheses are ranked using authority, evidence
strength, temporal consistency, independent corroboration, coverage, diversity,
and unresolved qualifications. Statistical co-occurrence or a high LLM
plausibility score cannot compensate for missing mandatory evidence.

After testing, deterministic verification marks each proposed claim
\textsc{Supported}, \textsc{Qualified}, \textsc{Contradicted}, or
\textsc{Unsupported}. Supported claims may appear normally; qualified claims
must state their limitations; contradicted claims may appear only as rejected
alternatives; and unsupported claims are removed. Numerical claims retain their
unit, entity, definition, period, and authoritative source, while explanatory claims
retain links to supporting evidence and evaluated alternatives. The final report
then states the observed condition, supported explanation, workload and system
context, important contradictions, limitations, and provenance.

\subsection{Trace-Localized and Regression-Gated Policy Evolution}
\label{sec:policy_evolution}

Every online execution records the interpreted scope, instantiated evidence requirements, source plan, tool events, normalized records, candidate hypotheses, claim--evidence links, final report, and completion state. When benchmark evaluation or reviewer feedback identifies a failure, an LLM-based critic uses this trace to localize the failure to scope interpretation, requirement construction, planning, retrieval, normalization, workload reasoning, verification, or composition.

The critic represents a proposed correction as a typed patch
\(\delta=\langle z,\omega,\kappa,d,\tau_s,f\rangle\), where \(z\) identifies the policy artifact, \(\omega\) is the update operation, \(\kappa\) defines the conditions under which the update applies, \(d\) is the patch payload, \(\tau_s\) identifies the supporting traces, and \(f\) identifies the expected failure family. Permitted targets are restricted to seven policy families: evidence requirements, source priorities and fallbacks, ontology mappings, temporal or entity rules, workload relations, normalization rules, and claim-verification rules.

For example, if the system produces a utilization explanation without examining queue evidence, the critic can propose a patch that makes queue state mandatory specifically for utilization-cause requests. The applicability condition prevents the change from affecting a direct request for an already reviewed utilization value. The critic cannot modify the active policy or deploy its proposal.

Before automated evaluation, a human acceptance decision \(H(\delta)\) checks whether the proposed patch is bounded, safe, and structurally valid. The human may approve or reject the patch but does not rewrite it. An approved candidate
\(\widehat{P}_{k+1}=P_k\oplus\delta\) is then evaluated on three groups: target cases associated with the observed failure, related cases that did not motivate the patch, and previously successful regression cases.

The automated release criteria were fixed before refinement. Let
\(\Delta Q_{\mathrm{target}}\) denote the proportion of previously failing target cases corrected by the candidate, \(\Delta Q_{\mathrm{heldout}}\) denote the change in held-out Pass@80 relative to the active policy, and \(\operatorname{RBR}\) denote the proportion of previously passing regression cases that fail under the candidate. Promotion is governed by

\begin{equation}
\begin{aligned}
G(\delta)={}&[H(\delta)=1]
\land[\Delta Q_{\mathrm{target}}\geq 0.10]
\land[\Delta Q_{\mathrm{heldout}}\geq 0] \\
&{}\land[\operatorname{RBR}\leq 0.10],\\
P_{k+1}={}&
\begin{cases}
\widehat{P}_{k+1}, & G(\delta)=1,\\
P_k, & \text{otherwise}.
\end{cases}
\end{aligned}
\label{eq:promotion-gate}
\end{equation}

Thus, an approved candidate must correct at least 10\% of the previously failing target cases, must not reduce held-out Pass@80, and may introduce an observed regression break rate of at most 10\%. Failure of any condition retains \(P_k\). Rejected candidates, evaluation outcomes, and failure reasons remain auditable, while promoted versions record their parent policy, patch, applicability condition, evaluation summary, and rollback target.

Equation~\eqref{eq:promotion-gate} is applied once for each agent in each of the five refinement rounds, producing 15 agent--round promotion opportunities. Ten candidates satisfy the complete gate and are promoted, while five fail at least one release condition and retain the previous policy, as summarized in Table~\ref{tab:policy-growth}. Figure~3(b) reports the corresponding round-level promotion counts and observed regression behavior. The zero final RBR reported for the gated strategy is an observed experimental outcome; it is not guaranteed by the gate, whose predefined maximum is 10\%.

The online policy therefore remains frozen during normal reporting, while the offline path treats each semantic change as a versioned and testable release. Policy growth is limited to typed operations over the seven permitted artifact families rather than unrestricted prompt or memory accumulation. RQ3 evaluates targeted correction, reuse on related held-out cases, promotion frequency, and protection of previously successful behavior within this environment.

\section{Evaluation}
\label{sec:evaluation}

We evaluate \textsc{REPORT-OPS} using the three research questions introduced
in Section~\ref{sec:introduction}. RQ1 measures whether claim-specific evidence
requirements improve report quality, claim support, and completion behavior. RQ2
measures whether typed records and explicit workload relations improve
operational explanations. RQ3 measures whether trace-localized policy updates
correct observed failures while preserving held-out and previously correct
behavior. To reduce dependence on a single generator or evaluator,
every configuration is evaluated with three agent models and
two separate LLM judge models.

\subsection{Benchmark and Ground Truth}
\label{sec:benchmark}

\paragraph{Base-scenario selection}
We construct the benchmark from a frozen, sanitized operational corpus from one HPC facility. The corpus contains reviewed periodic reports, utilization investigations, scheduler and queue observations, storage measurements, incident and maintenance records, and realistic staff-facing requests. These source families were selected because they participate in routine operational reporting but differ in authority, temporal granularity, and diagnostic purpose.

Candidate scenarios were retained when they represented a concrete staff-facing reporting task, identified a metric or operational condition that could be grounded in the corpus, and contained sufficient evidence to define the expected answer and completion behavior. We intentionally included both direct reporting tasks and cases requiring evidence from multiple source families. The benchmark is therefore a coverage-oriented, stratified collection rather than a random sample of the facility's request distribution.

The resulting 50 base scenarios cover availability (6), outage (7), downtime (4), utilization lookup or comparison (4), report metrics (7), analysis metrics or comparisons (6), report reasoning (10), and source capability (6). By reasoning complexity, 21 are Level~1 direct-reporting cases, 17 are Level~2 cross-source composition cases, and 12 are Level~3 diagnostic-discrimination cases.

\paragraph{Sanitization}
Sanitization removes personal identifiers, credentials, authentication material, internal access details, and conversational content unrelated to the reporting task. It preserves the public system names, metric values, dates, source roles, and temporal relationships needed to evaluate the answer. Free-text operational records are reduced to the evidence-bearing content needed by the reference rubric. The released query artifact contains the sanitized request, structured reference outcome, evidence metadata, and evaluation criteria; confidential source documents are not redistributed.

\paragraph{Controlled query expansion}
The 50 base scenarios are expanded into 68 benchmark queries by adding 18 controlled variants. Each variant changes an evidence condition or requested scope while preserving the underlying operational subject. This construction tests whether an agent responds to the validity of the evidence rather than merely recognizing the topic or retrieving a familiar record.

The resulting benchmark contains 48 complete-evidence queries, 4 missing-evidence queries, 4 temporally invalid queries, 3 authority-swapped queries, 3 contradictory-evidence queries, and 6 tool-failure queries. Their expected completion decisions are 55 \textsc{Answer}, 9 \textsc{Qualify}, and 4 \textsc{Block}. Each query specifies the requested system and reporting scope, mandatory and contextual evidence, acceptable source roles, required and prohibited claims, and expected completion behavior.

Missing-evidence variants remove a source family required for a complete conclusion. Temporal variants retain topically relevant evidence but move it outside the requested interval. Authority-swapped variants replace an authoritative record with a relevant but unsuitable source. Contradictory variants introduce evidence that challenges an otherwise plausible conclusion. Tool-failure variants distinguish an inaccessible source from a successful retrieval that contains no matching record.

Table~\ref{tab:benchmark-examples} illustrates the benchmark representation using two sanitized cases. The first permits a direct answer because the requested value, system, definition, and month are supported by an authoritative report. The second requires blocking the requested identity because the incident evidence identifies the type of faulty client but not a specific username.

\begin{table*}[t]
\centering
\caption{Sanitized benchmark examples showing the request, evidence state, requirements, and reference behavior.}
\label{tab:benchmark-examples}
\scriptsize
\setlength{\tabcolsep}{4pt}
\renewcommand{\arraystretch}{1.05}
\begin{tabularx}{\textwidth}{@{}p{0.14\textwidth}XX@{}}
\toprule
\textbf{Field} &
\textbf{Complete-evidence example} &
\textbf{Missing-evidence example} \\
\midrule
Request &
What was Perlmutter scheduled availability in October 2023? &
What exact NERSC username was behind the faulty client in the January 8--9, 2026 scratch outage? \\

Evidence &
A reviewed metrics report records Perlmutter scheduled availability as 98.3\% for October 2023. &
Reviewed incident evidence links the outage to a faulty accounting-service client but contains no username. \\

Requirement &
An authoritative value with matching system, metric definition, unit, and reporting month. &
An authoritative record explicitly identifying the requested user. Context identifying the client type is insufficient. \\

Expected state &
\textsc{Answer}. &
\textsc{Block}. \\

Reference behavior &
Report 98.3\%, including the system, metric, and month. &
State that the username cannot be identified from the available evidence; do not infer a person or account. \\
\bottomrule
\end{tabularx}
\end{table*}

\paragraph{Operational source roles}
The frozen corpus preserves the operational role of each source rather than flattening all records into a single relevance-ranked collection. Reviewed DOE metrics reports provide authoritative monthly values and approved commentary. Utilization investigations provide metric definitions, corrections, diagnostic hypotheses, and documented resolutions. Recurring monitoring sources provide queue depth, quality-of-service usage, account concentration, short-term workload pressure, storage read outliers, and write-performance baselines. Incident and maintenance records establish affected entities, event timelines, and impact intervals.

A separate source containing realistic staff requests, operational questions, and safety-testing interactions contributes evaluation cases but is not treated as authoritative operational evidence. Investigation records are reconstructed at the conversation or activity level so that corrections, rejected explanations, supporting evidence, and final conclusions remain connected to the original hypothesis. The reference annotations preserve these distinctions: diagnostic discussion may corroborate an explanation but cannot replace a reviewed historical metric, and a failed source call establishes evidence unavailability rather than the nonoccurrence of an event.

\paragraph{Explanatory subset}
RQ2 uses 13 explanatory queries: the 10 report-reasoning cases and three storage-anomaly cases drawn from the analysis-metric categories. Their annotations identify the observed condition, acceptable explanations, evidence required for each explanation, plausible but unsupported alternatives, and conditions requiring qualification. These cases require composition across multiple operational signals rather than retrieval of a single self-contained explanation.

Counterfactual consistency is evaluated through controlled changes to the evidence bundle, not physical interventions on the facility. Each comparison keeps the request and unrelated evidence fixed while changing one condition that should affect the explanation. Replacing a deep eligible queue and scheduler draining with a shallow queue should weaken a placement-constraint explanation. Moving an outage outside the requested interval should remove an outage-related explanation, and moving a CFS tail-latency spike outside an I/O-intensive workload's interval should remove a storage-impact explanation. A response is consistent when it changes the affected explanation without altering unrelated claims. These tests evaluate sensitivity to represented evidence relationships; they do not demonstrate unrestricted discovery of previously unknown causes.

\paragraph{Ground-truth construction}
\label{sec:ground-truth-construction}
The reference answers and evaluation rubrics were prepared from the frozen operational corpus and manually verified by the authors. Each annotation identifies the requested scope, required facts, supporting evidence, acceptable source roles, prohibited claims, required qualifications, and expected completion state.

A second author independently reviewed each annotation against its cited evidence, including factual values, entity and metric alignment, temporal scope, source authority, required qualifications, and completion behavior. Disagreements were documented and resolved through evidence-based adjudication: the reviewers jointly re-examined the relevant records and revised the annotation until the reference answer, evidence links, and evaluation criteria were mutually consistent. The study does not claim annotation by independent external HPC operators.

\paragraph{Coverage, scaling, and portability}
The 68-query benchmark provides controlled coverage of the reporting families and evidence failures studied here, but it is not intended to estimate their frequency in production. Its principal limitations are its size, its construction from one facility's corpus, and the overlap between the domain expertise used to define the benchmark and the expertise used to seed the reporting policy.

The benchmark can be extended by adding a grounded base scenario, specifying its evidence requirements and reference outcome, and applying the same missing, temporal, authority, contradiction, and tool-failure transformations where appropriate. This process does not require changing the evaluation logic, but every new case still requires evidence review and adjudication. Expert verification of the reference evidence, rather than automatic query generation, is therefore the primary scaling bottleneck.

The reporting method is not inherently limited to HPC, but its operational knowledge is domain-specific. Another HPC facility would need source adapters, local metric definitions, authority mappings, entity aliases, and facility-specific workload relations. A distributed-system deployment could use the same source-role, temporal-validation, contradiction, and answer/qualify/block mechanisms with service telemetry, deployment records, incident tickets, change logs, and service-level objectives. Such portability is architectural rather than empirically established: cross-facility and non-HPC transfer are not evaluated in this study.

\subsection{Experimental Setup}
\label{sec:experimental-setup}

\paragraph{Agent and judge matrix}
We evaluate three agent deployments:
\texttt{lbl/gpt-oss-20b},
\texttt{lbl/gemma-4-thinking}, and
\texttt{google/qwen-3}, denoted
\textsc{GPT-OSS-20B}, \textsc{Gemma-4-Thinking}, and
\textsc{Qwen-3}, respectively. These are the exact deployment identifiers used
by the evaluation endpoint; their prefixes identify serving routes rather than
model authorship. Documentation for the corresponding model families is
available from OpenAI, Google, and the Qwen project.\footnote{GPT-OSS-20B:
\url{https://developers.openai.com/api/docs/models/gpt-oss-20b};
Gemma 4 thinking:
\url{https://ai.google.dev/gemma/docs/capabilities/thinking};
Qwen3: \url{https://qwenlm.github.io/blog/qwen3/}.}

Each successfully generated report is evaluated separately by
\texttt{gemini-3.5-flash} and \texttt{mistral-large-3}.
Documentation for these judge-model families is provided by Google and
Mistral AI.\footnote{Gemini 3.5 Flash:
\url{https://ai.google.dev/gemini-api/docs/models/gemini-3.5-flash};
Mistral Large 3: \url{https://mistral.ai/news/mistral-3/}.}
The judges receive the sanitized request, reference rubric, expected completion
state, required and prohibited claims, qualification conditions, and generated
report. They do not receive the agent identity or the name of the evaluated
configuration.

For query \(q\), agent \(a\), configuration \(c\), and judge set
\(\mathcal{J}\), the report score is the arithmetic mean of the two judge
scores:

\begin{equation}
\overline{S}_{a,c}(q)
=
\frac{1}{|\mathcal{J}|}
\sum_{j\in\mathcal{J}} S_{a,c,j}(q),
\label{eq:multi-judge-score}
\end{equation}

where \(S_{a,c,j}(q)\in[0,100]\) and \(|\mathcal{J}|=2\). A report passes when
\(\overline{S}_{a,c}(q)\geq 80\). Unless otherwise stated, each metric is first
computed separately for each agent and then macro-averaged across the three
agents so that no agent contributes more weight because of differences in
output coverage.

\paragraph{Inference settings}
All agent and judge models were accessed through a common hosted inference
interface. For each configuration, the prompts, output limits, and execution
settings were held fixed across the evaluated queries. Runs used isolated
sessions without persistent memory. The complete evaluation configurations and
saved outputs are provided in the accompanying artifact.

\paragraph{Execution protocol}
RQ1 evaluates five configurations on all 68 queries. RQ2 evaluates four
reasoning configurations using identical evidence bundles for the 13
explanatory queries. RQ3 evaluates five update strategies over five sequential
target, held-out, and regression rounds.

Each final query--configuration--agent combination contributes at most one
report, and every included report receives evaluations from both judges.
Generation failures and model no-answers are recorded separately. Aggregate
results use the available scored reports, while paired statistical tests use
only query--agent pairs observed under both compared configurations. The
artifact contains the benchmark, evaluation configurations, saved outputs,
judgments, refinement partitions, and analysis code used to generate the
reported tables and figures.

\paragraph{Statistical analysis}
We compute 95\% confidence intervals using 4,000 nonparametric bootstrap
resamples \cite{efron1993bootstrap}. The implemented resampling unit is one
agent-level run for a configuration, with the query-level results already
aggregated within that run. The intervals therefore summarize variation across
the three evaluated agents and should be interpreted descriptively rather than
as population-level uncertainty over all possible operational queries.

For comparisons between configurations, continuous paired scores are tested
with the Wilcoxon signed-rank test
\cite{wilcoxon1945individual}, while paired Pass@80 outcomes are tested with
McNemar's test \cite{mcnemar1947note}. Both tests use matched query--agent
outputs. Reported \(p\)-values are two-sided and unadjusted for multiple
comparisons, so we use them as supporting evidence rather than as the sole
basis for a conclusion.

Judge reliability is summarized using Pass@80 agreement, Cohen's
\(\kappa\) \cite{cohen1960coefficient}, and the mean absolute difference
between the two judge scores. Confidence intervals, matched-pair counts, judge
coverage, and complete test outputs are provided in the artifact.

\subsection{Configurations and Metrics}
\label{sec:configurations}

\paragraph{RQ1: Evidence-guided reporting}
We compare five configurations. \textbf{Direct LLM} answers without retrieved
operational evidence. \textbf{Generic RAG} retrieves topically relevant records
without claim-specific requirements or source roles. \textbf{RAG with post-hoc
verification} checks an answer after generic retrieval. \textbf{REPORT-OPS
without evidence gating} uses source-aware planning and typed records but may
complete when mandatory evidence is missing or invalid. \textbf{Full
REPORT-OPS} additionally enforces evidence requirements, source-role and
temporal constraints, contradiction checks, and answer/qualify/block gating.

RQ1 reports \textbf{Pass@80}, the percentage of reports whose
two-judge mean is at least 80; completion accuracy, the
percentage of queries for which the agent's structured
\textsc{Answer}, \textsc{Qualify}, or \textsc{Block}
decision matches the reference decision; State F1, macro F1
over the three completion decisions; false completion rate
(FCR), the percentage of cases requiring qualification or
blocking that instead receive an unqualified answer; and
claim-support precision (CSP), the percentage of assessed
factual and explanatory claims supported by valid evidence or stated
as permitted qualified inferences. Source-role,
temporal-consistency, and contradiction-handling accuracy are
retained as diagnostic measures.

\paragraph{RQ2: Workload-aware operational explanation}
All four configurations receive the same underlying evidence bundle.
\textbf{Raw-evidence reasoning} operates on unstructured evidence;
\textbf{typed records} creates structured metric, event, status, workload,
storage, and commentary records; \textbf{typed records with workload
relations} adds relations among workload behavior, queue pressure, utilization,
incidents, maintenance, and storage; and \textbf{full claim verification}
further checks required evidence, source roles, temporal overlap,
contradictions, missing evidence, and rejected hypotheses.

RQ2 reports Pass@80, \textbf{explanation accuracy},
\textbf{counterfactual consistency (CFCons)}, and
\textbf{rejected-hypothesis accuracy (RHA)}. Explanation accuracy requires an
explanation supported by the reference evidence without asserting a prohibited
cause. CFCons tests whether an explanation changes in the expected direction
when one relevant condition is altered while the remaining evidence is held
fixed. RHA measures the percentage of reference-rejected hypotheses correctly
ruled out. These metrics evaluate evidence-grounded hypothesis testing, not
unrestricted discovery of previously unknown physical causes.

\paragraph{RQ3: Human-supervised policy refinement}
We compare \textbf{no refinement}, \textbf{final-answer refinement},
\textbf{global prompt update}, \textbf{localized update without a regression
gate}, and \textbf{localized update with a regression gate}. The localized methods may
change a failure-specific evidence requirement, source policy, ontology
relation, temporal rule, workload relation, or verification rule; the gated
method promotes the change only after target, held-out, and regression tests.

RQ3 reports final Pass@80, \textbf{targeted fix rate (TFR)},
\textbf{$\Delta$ held-out}, and \textbf{regression break rate (RBR)}. TFR is the
percentage of initially failed target cases corrected after refinement.
$\Delta$ held-out is the percentage-point change on
round-specific cases held out from the motivating target set. RBR is the percentage of initially passing regression cases
that fail in a later refinement round. Under the no-refinement
control, it captures run-to-run model variability; under the
update strategies, it may additionally capture policy-induced
regressions. Lower is better. All values are macro-averaged across agents.

\subsection{Refinement Protocol}
\label{sec:refinement-protocol}

RQ3 is conducted independently for each agent over five rounds. In each round,
failed target traces motivate a bounded candidate update, held-out cases test
whether the change generalizes, and regression cases test whether previously
correct behavior is preserved. Target and held-out cases are disjoint, whereas
regression cases may recur across rounds to provide continuing protection
against behavioral breakage. The experiment therefore uses round-specific
partitions rather than one globally disjoint three-way split.

Each agent begins from the same initial policy, and updates learned for one
agent are not transferred to another. The agent generates candidate repair
patches from prior-run failures. A human acceptance gate may approve or reject a
candidate for safety and artifact integrity, but authors do not rewrite it.
Accepted candidates in the gated strategy are promoted only after predefined
target, held-out, and regression criteria pass; rejected candidates leave the
active policy unchanged. The localized strategy ``without a regression gate''
retains the same non-editing human acceptance check but omits the automated
held-out and regression release gate. The process is therefore human-supervised policy refinement, not autonomous
self-modification. A patch demonstrates reuse only when it improves cases that
did not motivate the update. The reported held-out gain therefore measures
within-environment, same-family generalization; updates are not transferred
between agents or facilities. Exact partitions, typed patches, decisions, and
promotion history are included in the artifact.

\begin{table}[t]
\centering
\caption{Observed policy evolution over five rounds.}
\label{tab:policy-growth}
\scriptsize
\setlength{\tabcolsep}{3pt}
\renewcommand{\arraystretch}{0.90}
\begin{tabularx}{\columnwidth}{@{}Xr@{}}
\toprule
\textbf{Refinement property} & \textbf{Observed value} \\
\midrule
Sequential rounds / agent models & 5 / 3 \\
Agent--round promotion opportunities & 15 \\
Promoted updates across rounds & 10 \\
Previous policy retained & 5 \\
Permitted typed patch families & 7 \\
Human edits to patch contents & 0 \\
Cross-agent or cross-facility transfer & Not evaluated \\
\bottomrule
\end{tabularx}
\end{table}

\subsection{Results}
\label{sec:results}

Table~\ref{tab:main-results} reports results macro-averaged across the three
agent models. The configurations are cumulative ablations: they measure the
combined effect of LLM reasoning, retrieved evidence, author-seeded operational
knowledge, and automated verification. They do not measure the LLM's reasoning
ability independently of the policy supplied to it.

\begin{table*}[!hbt]
\centering
\caption{Principal evaluation results. Arrows indicate the preferred
direction for each metric. Bold denotes the best value in each column;
ties are all bolded. RQ2 is evaluated on the 13 explanatory queries.}
\label{tab:main-results}

\footnotesize
\setlength{\tabcolsep}{6.0pt}
\renewcommand{\arraystretch}{0.80}

\begin{tabularx}{0.85\textwidth}{
@{}
>{\raggedright\arraybackslash}X
S[table-format=2.2,detect-weight=true]
S[table-format=2.2,detect-weight=true]
S[table-format=-2.2,detect-weight=true]
S[table-format=3.2,detect-weight=true]
@{}
}
\toprule

\multicolumn{5}{@{}l}{
\textbf{RQ1: Evidence validity and selective completion}
}\\[-1pt]

\textbf{Configuration}
&
\multicolumn{1}{c}{
\makecell{\textbf{Pass@80}\\[-1pt]\(\uparrow\)}
}
&
\multicolumn{1}{c}{
\makecell{\textbf{State F1}\\[-1pt]\(\uparrow\)}
}
&
\multicolumn{1}{c}{
\makecell{\textbf{FCR}\\[-1pt]\(\downarrow\)}
}
&
\multicolumn{1}{c}{
\makecell{\textbf{CSP}\\[-1pt]\(\uparrow\)}
}
\\

\midrule

Direct LLM
& 4.43
& \textbf{45.06}
& \textbf{75.99}
& 24.86
\\

Generic RAG
& 6.43
& 27.46
& 97.44
& 23.55
\\

RAG + verification
& 7.02
& 30.55
& 89.75
& 25.61
\\

REPORT-OPS, no evidence gate
& \textbf{70.42}
& 32.86
& 82.05
& 40.58
\\

Full REPORT-OPS
& 68.63
& 33.58
& 76.92
& \textbf{90.06}
\\

\addlinespace[3pt]
\midrule
\addlinespace[1pt]

\multicolumn{5}{@{}l}{
\textbf{RQ2: Evidence-grounded operational explanation}
}\\[-1pt]

\textbf{Configuration}
&
\multicolumn{1}{c}{
\makecell{\textbf{Pass@80}\\[-1pt]\(\uparrow\)}
}
&
\multicolumn{1}{c}{
\makecell{\textbf{Expl. Acc.}\\[-1pt]\(\uparrow\)}
}
&
\multicolumn{1}{c}{
\makecell{\textbf{CFCons}\\[-1pt]\(\uparrow\)}
}
&
\multicolumn{1}{c}{
\makecell{\textbf{RHA}\\[-1pt]\(\uparrow\)}
}
\\

\midrule

Raw evidence
& 0.00
& 30.53
& 31.39
& \textbf{100.00}
\\

Typed records
& 0.00
& 33.95
& 33.89
& 98.72
\\

\hspace{0.75em}+ workload relations
& 33.97
& 51.37
& \textbf{41.18}
& \textbf{100.00}
\\

\hspace{0.75em}+ claim verification
& \textbf{41.02}
& \textbf{55.54}
& 38.33
& \textbf{100.00}
\\

\addlinespace[3pt]
\midrule
\addlinespace[1pt]

\multicolumn{5}{@{}l}{
\textbf{RQ3: Human-supervised policy refinement}
}\\[-1pt]

\textbf{Update strategy}
&
\multicolumn{1}{c}{
\makecell{\textbf{Final Pass@80}\\[-1pt]\(\uparrow\)}
}
&
\multicolumn{1}{c}{
\makecell{\textbf{TFR}\\[-1pt]\(\uparrow\)}
}
&
\multicolumn{1}{c}{
\makecell{\(\boldsymbol{\Delta}\) \textbf{Held-out}\\[-1pt]\(\uparrow\)}
}
&
\multicolumn{1}{c}{
\makecell{\textbf{RBR}\\[-1pt]\(\downarrow\)}
}
\\

\midrule

No refinement
& 44.44
& 2.56
& 0.00
& 3.03
\\

Answer refinement
& 85.19
& 78.63
& -1.33
& 9.09
\\

Global prompt update
& 90.74
& 87.47
& 6.67
& 9.76
\\

Localized, no regression gate
& \textbf{94.44}
& \textbf{91.17}
& \textbf{9.33}
& 6.06
\\

Localized + regression gate
& \textbf{94.44}
& \textbf{91.17}
& \textbf{9.33}
& \textbf{0.00}
\\

\bottomrule
\end{tabularx}
\end{table*}

\begin{figure*}[!t]
\centering
\vspace{-2mm}
\begin{minipage}[t]{0.485\textwidth}
\centering
\includegraphics[width=\linewidth]{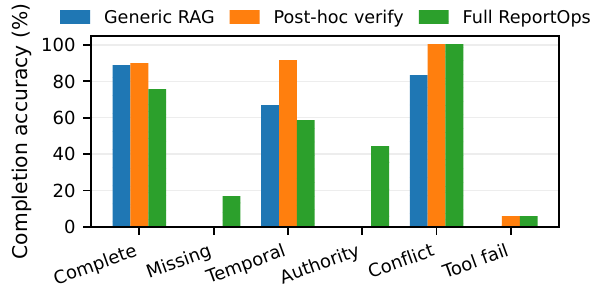}
\vspace{-1.5mm}

{\scriptsize\textbf{(a) Completion accuracy by evidence state}}
\end{minipage}
\hfill
\begin{minipage}[t]{0.485\textwidth}
\centering
\includegraphics[width=\linewidth]{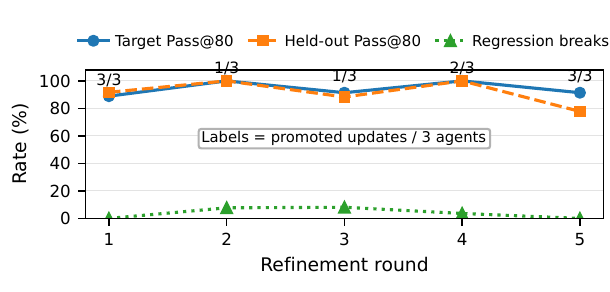}
\vspace{-1.5mm}

{\scriptsize\textbf{(b) Quality and regressions by refinement round}}
\end{minipage}
\vspace{-1mm}
\caption{Evaluation summary across three agent models.
Panel (a) shows completion accuracy under controlled evidence
states. Panel (b) shows target-set and round-held-out Pass@80,
regression breaks, and agent-specific promotions after the human acceptance and
automated release gates; labels such as $3/3$ indicate promotion for all three
agent models.}
\label{fig:evaluation-summary}
\vspace{-2mm}
\end{figure*}

\paragraph{RQ1: Evidence validity and selective completion}
Full \textsc{REPORT-OPS} achieves 68.63\% Pass@80, compared with 7.02\% for the
strongest retrieval baseline. Its improvements over Direct LLM, Generic RAG,
and RAG with post-hoc verification are significant under both paired tests
(\(p<0.001\)). Generic retrieval and post-hoc answer checking provide only
small gains over Direct LLM. The large Pass@80 increase appears when
source-aware planning, typed records, and claim-specific operational
requirements are introduced in the ungated \textsc{REPORT-OPS} configuration.
This result reflects the combination of LLM reasoning with explicit,
author-seeded domain structure; it should not be interpreted as an improvement
produced by the LLM alone.

Adding the evidence gate does not increase Pass@80 relative to ungated
\textsc{REPORT-OPS} (68.63\% versus 70.42\%), and the difference is not
statistically significant. Its principal benefit is evidence discipline: CSP
increases from 40.58\% to 90.06\%. Figure~\ref{fig:evaluation-summary}(a) shows
that the gate is particularly useful when required evidence is missing or comes
from an inappropriate source.

Full \textsc{REPORT-OPS} does not dominate every RQ1 metric. Direct LLM obtains
the highest State F1 and a slightly lower FCR, while all configurations retain
substantial difficulty selecting the exact \textsc{Answer},
\textsc{Qualify}, or \textsc{Block} state. Performance is also weaker under
temporally invalid evidence. RQ1 therefore supports two narrower conclusions:
source-aware, structured planning produces the main report-quality gain, and
deterministic evidence gating substantially improves claim-support precision.
Exact completion-state selection remains an open problem.

\paragraph{RQ2: Evidence-grounded operational explanation}
Converting raw evidence into typed records yields only a small increase in
explanation accuracy, from 30.53\% to 33.95\%, and neither configuration
produces passing explanatory reports. The principal improvement appears after
adding the author-seeded workload relations. Pass@80 reaches 33.97\%,
explanation accuracy reaches 51.37\%, and CFCons reaches its highest value of
41.18\%. These results show that the LLM benefits from explicit relations among
workload behavior, queue pressure, utilization, scheduler conditions,
incidents, maintenance, and storage observations. They do not show that the LLM
independently discovers those relations.

Full claim verification obtains the highest Pass@80 and explanation accuracy,
41.02\% and 55.54\%, but its improvement over workload relations alone is not
statistically significant. It also reduces CFCons from 41.18\% to 38.33\%.
RHA is approximately saturated across the configurations and therefore provides
little discrimination.

RQ2 contains only 13 explanatory queries, and its counterfactual tests modify
represented queue, outage, and storage evidence rather than introducing
previously unknown failure mechanisms. The results consequently support
evidence-grounded hypothesis testing over the encoded relation families, not
open-ended causal discovery. The relatively low absolute Pass@80 and
explanation-accuracy values further show that reliable operational explanation
remains harder than factual reporting.

\paragraph{RQ3: Human-supervised policy refinement}
Localized refinement provides the strongest overall result. Both localized
strategies reach 94.44\% final Pass@80, correct 91.17\% of targeted failures,
and improve performance on related held-out cases by 9.33 percentage points.
Without the regression gate, localized refinement produces a 6.06\% observed
regression break rate. The gated strategy retains the same target and held-out
results while reducing the final observed RBR to zero.

Figure~\ref{fig:evaluation-summary}(b) shows that the release gate does not
promote every candidate that improves its motivating cases. Across 15
agent--round opportunities, 10 candidate updates are promoted and five retain
the previous policy, as summarized in Table~\ref{tab:policy-growth}. Answer
refinement and global prompt updates also improve target performance, but
produce weaker held-out behavior or more observed regressions. The 3.03\% RBR
under no refinement additionally indicates that some measured breakage can
arise from model variability rather than policy changes alone.

The zero RBR of the gated strategy is an observed outcome of these five rounds,
not a guarantee that future updates cannot cause regressions. The 9.33-point
held-out improvement demonstrates reuse on related cases within the same
environment. It does not establish transfer between agents, HPC facilities, or
previously unseen operational domains.

\paragraph{Judge agreement}
The two judges agree on more than 96\% of Pass@80 decisions in both RQ1 and
RQ2, with Cohen's \(\kappa\) between 0.87 and 0.89. The thresholded conclusions
are therefore consistent across the two evaluated judges. Their continuous
scores differ by approximately ten points on average, however, showing that
the choice of judge still affects score magnitude. Agreement between two LLM
judges does not substitute for evaluation by independent HPC operators.

\subsection{Threats to Validity}
\label{sec:limitations}

\paragraph{Benchmark and policy alignment}
The benchmark, reference rubrics, and initial operational policy are derived
from one sanitized HPC corpus. The same domain expertise therefore influences
both the expected answers and the source roles, evidence requirements, and
workload relations used by \textsc{REPORT-OPS}. This creates possible
corpus-specific alignment. The staged ablations expose the contributions of
retrieval, typed representation, workload relations, evidence gating, and claim
verification, but they do not completely separate LLM capability from manually
authored domain knowledge.

\paragraph{Scale and external validity}
The benchmark contains 68 purposefully selected queries and is not a random
sample of production requests. RQ2 is especially limited, with 13 explanatory
queries drawn from represented explanation families. All evaluations use one
facility environment. We therefore do not claim that the measured gains
transfer unchanged to another HPC facility or to a non-HPC distributed system.
Such deployments would require new source adapters, metric definitions,
authority mappings, evidence templates, and operational relations.

\paragraph{Ground truth and judging}
Reference annotations were prepared by the authors, independently reviewed by a
second author, and adjudicated against the cited evidence. They were not
annotated by independent external HPC operators. The two LLM judges receive the
same reference rubric and may share model or prompt biases. Their high Pass@80
agreement supports consistency between the selected evaluators, but does not
establish human-review agreement or operational usefulness in deployment.

\paragraph{Statistical validity}
Each final query--configuration--agent combination contributes at most one
scored output, so residual model variability is not averaged over repeated
generations. Some model no-answers and execution failures produce unequal
denominators; paired tests therefore use only jointly observed query--agent
outputs. Confidence intervals summarize variation across three agent-level
runs, and reported \(p\)-values are unadjusted for multiple comparisons. These
statistics should consequently be interpreted as supporting evidence for the
observed comparisons rather than precise population-level estimates.

\paragraph{Refinement validity}
RQ3 evaluates five rounds of refinement within the same environment.
Round-specific held-out cases test reuse beyond the cases that motivate each
patch, while regression cases test previously successful behavior. These sets
remain smaller and more controlled than a long-running production workload.
The experiment does not evaluate cross-agent patch transfer, cross-facility
transfer, indefinite policy growth, or long-term overfitting to the regression
suite. The frozen online policy, typed patches, non-editing human acceptance
step, version history, and automated release gate make the evaluated process
inspectable, but longer-term operator deployment remains future work.

\section{Conclusion}
\label{sec:conclusion}

This paper presented \textsc{REPORT-OPS}, a hybrid agent for evidence-grounded HPC operational reporting. The LLM interprets requests, plans evidence collection, proposes explanations, links claims to records, and drafts reports. Author-seeded, versioned policy defines source roles, evidence requirements, and operational relations, while automated checks enforce entity, metric, temporal, contradiction, and completion constraints. The evaluation shows that the principal report-quality improvement appears when LLM reasoning is combined with source-aware planning and structured operational knowledge. Evidence gating does not increase raw Pass@80, but substantially improves claim-support precision by preventing unsupported or insufficiently qualified claims.

For operational explanation, explicit workload relations improve performance over raw evidence and typed records alone. These results support evidence-grounded testing of represented explanation families, not unrestricted causal discovery. The 13-query explanatory evaluation remains limited, and the results do not establish that the LLM can independently discover previously unseen operational causes.

\textsc{REPORT-OPS} also supports human-supervised policy refinement. An LLM critic proposes bounded, trace-localized policy changes; humans approve or reject the unchanged proposals; and automated target, held-out, and regression criteria control promotion. Across five rounds, localized refinement improves targeted and related held-out cases, while the gated strategy introduces no observed final regression breaks. This demonstrates reuse within the evaluated environment, not guaranteed safety, cross-agent learning, cross-facility transfer, or indefinite resistance to regression-suite overfitting.

The framework is intended to assist rather than replace HPC operators. Its evidence-validation and release-control architecture may be adapted to other facilities or operational domains, but its source adapters, metric definitions, authority mappings, evidence templates, and operational relations require domain-specific configuration. Future work will evaluate independently constructed and cross-facility benchmarks, automatically induced policy artifacts, previously unseen explanation families, external operator review, and longer-term policy evolution.





\section*{Acknowledgments}
This research was supported by the National Energy Research Scientific
Computing Center (NERSC), a U.S. Department of Energy Office of Science
User Facility operated under Contract No.~DE-AC02-05CH11231.

\bibliographystyle{IEEEtran}
\bibliography{references}
\end{document}